\documentclass[preprint, nonatbib]{article} 

\pdfoutput=1

\usepackage{neurips_2023}

\usepackage[utf8]{inputenc} 
\usepackage[T1]{fontenc}    
\usepackage[hidelinks]{hyperref}       
\usepackage{url}            
\usepackage{booktabs}       
\usepackage{amsfonts}       
\usepackage{nicefrac}       
\usepackage{microtype}      
\usepackage{xcolor}         
\usepackage{graphicx}
\usepackage{subcaption}
\usepackage{amsmath}
\usepackage{bbm}
\usepackage[linesnumbered,ruled]{algorithm2e}

\newcommand{\bfx}{\mathbf{x}}
\newcommand{\bfy}{\mathbf{y}}
\newcommand{\bfz}{\mathbf{z}}
\newcommand{\bfe}{\mathbf{e}}

\newcommand{\bbX}{\mathbb{X}}
\newcommand{\bbY}{\mathbb{Y}}
\newcommand{\bbZ}{\mathbb{Z}}
\newcommand{\bbR}{\mathbb{R}}

\newcommand{\kmax}{k_\mathrm{max}}

\title{Information-Ordered Bottlenecks for Adaptive Semantic Compression}

\author{%
  Matthew Ho$^1$\\
  \texttt{\href{mailto:matthew.annam.ho@gmail.com}{matthew.annam.ho@gmail.com}}\\
  \And 
  Xiaosheng Zhao$^{1,2}$\\
  \texttt{\href{mailto:xszhao20@gmail.com}{xszhao20@gmail.com}}\\
  \And 
  Benjamin Wandelt$^{1,3,4}$\\
  \texttt{\href{mailto:benwandelt@gmail.com}{benwandelt@gmail.com}}\\
  \And
  \vspace{-1em}\\
  $^1$Institut d'Astrophysique de Paris, CNRS \& Sorbonne Universit\'{e}\\ 
  $^2$Department of Astronomy, Tsinghua University\\ 
  $^3$Center for Computational Astrophysics, Flatiron Institute\\ 
  $^4$Sorbonne Universit\'{e}, Institut Lagrange de Paris 
}

\begin{document}

\maketitle

\begin{abstract}
We present the information-ordered bottleneck (IOB), a neural layer designed to adaptively compress data into latent variables ordered by likelihood maximization. Without retraining, IOB nodes can be truncated at any bottleneck width, capturing the most crucial information in the first latent variables. Unifying several previous approaches, we show that IOBs achieve near-optimal compression for a given encoding architecture and can assign ordering to latent signals in a manner that is semantically meaningful. IOBs demonstrate a remarkable ability to compress embeddings of image and text data, leveraging the performance of SOTA architectures such as CNNs, transformers, and diffusion models. Moreover, we introduce a novel theory for estimating global intrinsic dimensionality with IOBs and show that they recover SOTA dimensionality estimates for complex synthetic data. Furthermore, we showcase the utility of these models for exploratory analysis through applications on heterogeneous datasets, enabling computer-aided discovery of dataset complexity. 
\end{abstract}

\section{Introduction}

Modern deep neural networks (DNNs) \cite{lecun2015deep} excel at discovering complex signals and constructing informative high-level latent representations. However, their complexity often leads to computational burden, memory requirements, lack of interpretability, and potential overfitting. Latent compression, which reduces the dimensionality of a network's latent space while preserving essential information, has emerged as a solution to these challenges. In this paper, we propose a generic method for adaptively compressing and ordering the latent space of any DNN architecture, with applications in data exploration and model interpretability.

Classical linear methods, such as Principle Component Analysis (PCA) \cite{jolliffe2016principal}, fail when dealing with nonlinearly correlated features in datasets. Nonlinear extensions like kernel PCA \cite{scholkopf2005kernel} offer some improvements, but they often struggle or become intractable in high-dimensional data. However, deep autoencoders have demonstrated remarkable capabilities in fitting nonlinear manifolds, even in high-dimensional settings \cite{kingma2013auto, gondara2016medical, he2022masked}. Despite their frequentist training procedure, these autoencoders have theoretical foundations providing Bayesian guarantees on their expressibility \cite{tishby1999, alemi2016deep, saxe2019information}. Moreover, the rise of multimodal models and zero-shot inference \cite{ramesh2021zero, ramesh2022hierarchical} has sparked interest in creating and understanding semantic latent spaces in DNNs \cite{radford2021learning}.

The goal of our proposed approach is to learn to embed latent information hierachically such that, at inference time, we can dynamically vary the bottlenck width while ensuring likelihood maximization. We achieve this by providing a loss-based incentive for our neural network to prioritize certain neural pathways over others. The secondary effects of this are that we can study the inference performance of our model as a function of bottleneck width and interpret which data features carry the most information. In short, the contributions of this work are as follows:
\begin{itemize}
    \item We introduce the information-ordered bottleneck (IOB), a neural layer that compresses high-level data features and orders them based on their impact on likelihood maximization, unifying previous approaches under a single framework.
    \item Through autoencoding experiments on synthetic datasets, our IOB models achieve optimal data compression for a given neural architecture while capturing intrinsic semantic properties of the data distribution.
    \item Applying our method to the high-dimensional CLIP semantic embedding space, IOBs effectively capture and organize latent information from state-of-the-art (SOTA) image and text encoders.
    \item We propose a novel methodology for estimating intrinsic dimensionality using IOBs, achieving SOTA performance on high-dimensional synthetic experiments.
\end{itemize}



\section{Related work}
The authors of Nested Dropout \cite{rippel2014learning} showed that randomly truncating the width of a hidden layer during training could encourage a latent ordering in autoencoders. They showed that, in the case of linear activation functions, Nested Dropout exactly recovers the PCA solution. They later explored applications of this method for data compression and resource management and subsequent authors have applied it in the context of convolutional layers \cite{finn2014learning}, normalizing flows \cite{bekasov2020ordering}, and fully nested networks \cite{cui2020fully}.
Independently, the authors of \cite{staley2022triangular} introduced Triangular Dropout, which deterministically weighted the loss function of an autoencoder using reconstructions from every possible configuration of truncated hidden layers. This marginalization over truncation was computed exactly, at every training step, instead of stochastically as in \cite{rippel2014learning}.
In subsequent sections, we show that these two methods are special cases of IOBs, subject to certain hyperparameter choices, and compare them empirically under the same experiments.

The Principle Component Analysis Autoencoder (PCAAE) \cite{pham2022pca} was introduced as an ordered encoder-decoder framework which enforced latent disentanglement, i.e. explicitly penalizing non-independent latent variables. This method computes latent embeddings by learning one latent variable at a time with separate encoders. Concurrent with each new latent encoder, a separate decoder of new width is trained. As a result, the training procedure complexity is equivalent to that of training $\kmax$ separate encoders, where $\kmax$ is the maximum bottleneck width. \cite{staley2022triangular} showed that this method results in notably worse compression than Triangular Dropout on MNIST. Due to this and its significant computational expense for high-dimensional embeddings, this method was excluded from our analysis.

\section{Information-Ordered Bottlenecks} \label{sec:theory}
Consider a dataset of $N$ input-output pairs $\mathcal{D}:= \{(\bfx_i, \bfy_i)\}_{i=1}^N$ for inputs $\bfx \in \bbX$ and outputs $\bfy\in \bbY$. We assume there exists an optimal relationship $f^*:\bbX\rightarrow\bbY$ which maximizes a given joint log-likelihood $\ell:(\bbX,\bbY)\rightarrow \mathbb{R}$ for all $\bfx\in\bbX$ and $\bfy \in \bbY$. In this application, we attempt to learn a mapping $\bbX\rightarrow\bbY$ through the composition of two parametric transformations $e_\phi : \bbX\rightarrow \bbZ$ and $d_\eta : \bbZ \rightarrow \bbY$, which respectively map inputs $\bfx$ to latent representations $\bfz\in \bbZ$ and, subsequently, latent representations $\bfz$ to outputs $\bfy$. We define the composition of these functions as $f_\theta := d_\eta \circ e_\phi$ described by a joint parameter vector $\theta := \{\phi,\eta\}$. In this context, our goal is then to compute:
\begin{equation} \label{eqn:fstar}
  \hat\theta = \arg\max_{\theta\in\Theta}\sum_{i=1}^N\ell\left[f_\theta\left(\bfx_i\right), \bfy_i\right].
\end{equation}
This formulation is generalizable to many machine learning problems, including regression, classification, and autoencoding or variational density estimation.

This problem is considered a bottleneck if the dimensionality of the latent representation space is smaller than that of either the input or output space. i.e. $\dim(\bbZ)<\dim(\bbX)$ or $\dim(\bbZ)<\dim(\bbY)$. Under this condition, if $e_\phi$ and $d_\eta$ are linear transformations, then their composition $f_\theta$ is not sufficiently flexible to model every arbitrary $f^*$. In other words, it is impossible to guarantee linear embedding of high-dimensional information into low-dimensional spaces without information loss. However, for sufficiently flexible non-linear $e_\phi$ and $d_\eta$, such as deep neural networks, it is theoretically possible to compress any distribution in $\bbX$ into $\bbZ$ where $0<\dim(\bbZ)<\dim(\bbX)$. In fact, both analytic and empirical evidence suggests that the introduction of a bottleneck actually improves the training procedure of complex problems \cite{saxe2019information}. However, the dimensionality of $\bbZ$ is often taken to be a hyperparameter, with trade offs for fitting accuracy versus constraining power. For the remainder of this work, we will consider $e_\phi$ and $d_\eta$ to be deep neural networks where $\phi$ and $\eta$ are the tunable weights and biases of each neural layer.

\begin{figure}[!t]
  \centering
    \includegraphics[width=0.75\textwidth]{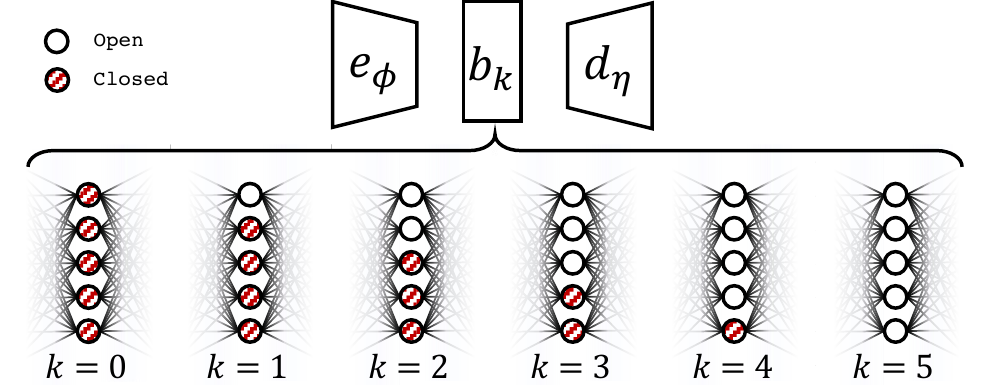}
    \caption{Example concept diagram of IOBs where $\kmax=5$. During each training step, the bottleneck width $k$ is varied through masking. The model is optimized with the summative loss in Equation \ref{eqn:objective}. This results in a model which is incentivized to pass as much information as possible through the top nodes, which are the most reliable during training.}
    \label{fig:concept}
\end{figure}

The key concept of our approach is to implement a training loss which maximizes Equation \ref{eqn:fstar} for all possible bottleneck dimensionalities. First, we introduce an IOB layer $b_k:\bbZ\rightarrow\bbZ$ designed to dynamically mask out information flowing through our latent representations $\bfz$ to form an effective bottleneck of dimensionality $k$. The concept is represented graphically in Figure \ref{fig:concept}. For a given $k$, $b_k$ functions by assigning the output of every $i$-th node equal to 0 for all $i>k$. These `closed' nodes thus can neither pass any information on to subsequent layers nor propagate gradients back to previous layers. We note that the order in which we open nodes with increasing $k$ is fixed, such that if the $i$-th node is open, then all nodes $j<i$ must also be open. Next, we use this IOB $b_k$ to construct a bottlenecked neural network, using our $e_\phi$ and $d_\eta$. The composition of these three transformations, $f^{(k)}_\theta := d_\eta \circ b_k \circ e_\phi:\bbX\rightarrow\bbY$, is now an input-output mapping with an adjustable latent dimensionality $\dim(\bbZ) = k$.
Finally, we introduce an extension of Equation $\ref{eqn:fstar}$ which maximizes the log-likelihood over all bottlenecks:
\begin{equation}\label{eqn:objective}
  \hat\theta = \arg\max_{\theta\in\Theta}\sum_{i=1}^N\sum_{k=0}^{k_\mathrm{max}}\rho_k\ell\left[f^{(k)}_\theta\left(\bfx_i\right), \bfy_i\right].
\end{equation}
This new objective function is now a linear combination of the log-likelihood of the model with each bottleneck size $k$, weighted by scalars $\rho_k$ and taken up to a maximum $\kmax$, which can be set as $\kmax = \min(\dim(\bbX), \dim(\bbY))$ without loss of generality.


The choices of $\rho_k$'s and the method for computing the summation over $k$ in Equation \ref{eqn:objective} are hyperparameters that allow us to generalize this framework to previous approaches. An example of this framework can be seen in the implementation of Nested Dropout \cite{rippel2014learning}, where $\rho_k$ values follow a geometric distribution with rate $r<1$. The summation over $k$ is stochastically calculated in each training batch. To address the vanishing gradient problem observed at high $k$, the authors of Nested Dropout introduced a `unit-sweeping' procedure. This procedure freezes the learned latents at low $k$ after convergence, leveraging the memoryless property of the geometric distribution with low rate $r<<1$ to reduce the optimization's condition number. Another implementation, Triangular Dropout \cite{staley2022triangular}, uses a constant $\rho_k$ for all $k$ and computes the sum over $k$ exactly in each training step. We implement and compare these methods, examining their properties in practical applications.

The motivation behind this design is to train a neural network to simultaneously learn signal features from data and compress the most important ones into as low-dimensional space as it can manage. In Figure \ref{fig:concept}, the top $k=1$ latent variable is the most reliable pathway, as it is open in $k-1/k$ terms of the loss function. This encourages the network to prioritize passing information through the top node, leveraging architectural flexibility. The resulting latent layer exhibits an ordered arrangement of components based on their contribution to the log-likelihood, with more informative nodes at low $k$. During inference, incremental opening of the bottleneck allows for improvements in signal modeling and gains in the log-likelihood. This approach supports model interpretability (Section \ref{sec:exploration}) and intrinsic dimensionality analysis in autoencoding (Section \ref{sec:dimensionality}). This adaptive compressor design does not enforce a fixed bottleneck width or strict latent disentanglement \cite{pham2022pca}. Instead, these properties are learned directly from the data, motivating the neural network to efficiently compress the data by maximizing gains in the log-likelihood through low $k$.



\subsection{Implementation}
We implement two forms of IOBs, heretoafter referred to as Linear IOB and Geometric IOB. The former uses a constant weighting in Equation \ref{eqn:objective}, i.e. $\rho_0 = \rho_k\ \forall\ k$. The latter uses a geometric distribution of $\rho_k$ such that $\rho_k = (1-r)^kr$ where $r$ is a hyperparameter. This hyperparameter varies dependent on the experiment and is listed in Appendix \ref{apx:implementation}. In addition, for the Geometric IOB we implement the stochastic loss summation and unit sweeping procedure, following \cite{rippel2014learning}. We found that, without this procedure, the vanishing gradient problem due to the geometric weighting is too strong to make any training progress.

\section{Experiments} \label{sec:experiments}
We describe several synthetic and real datasets which are used throughout this work to experiment with IOBs. All models implemented in this study use an autoencoder setting, i.e. with $\bbY=\bbX$, though we suggest extensions to further inference problems as future work. Additional implementation details (e.g. specific architectures, training procedures, optimizers, etc.) are provided in Appendix \ref{apx:implementation}.

\subsection{S-curve}
The first dataset is a commonly-used toy example for manifold learning. We sample a set of $N=10,000$ data points from a noisy S-curve distribution in $\bbR^3$. The S-curve distribution is a curved 2D manifold in 3D space. We also add a small amount of isotropic Gaussian noise to each sample with a standard deviation $0.1$. For reproducibility, we use the $\texttt{sklearn.datasets}$ implementation of the S-curve distribution to sample these points. The 3D and projected distributions of these sampled datapoints are shown in Figure \ref{fig:ex_s}.
The autoencoder architecture for this dataset is a feed-forward neural network with three dense layers in each of the encoder $e_\phi$ and decoder $d_\eta$ and a bottleneck of max width $\kmax=4$. 

\subsection{$n$-Disk}
We next define a series of complex synthetic datasets for which the intrinsic dimensionalities are known a priori. We define a generator which produces single-channel images of size $32\times32$. In short, the generator places one or several `disks' in an otherwise empty image by changing the pixel values of circular regions from zero to one. Each disk is allowed to overlap with others. For a termed $n$-Disk dataset, we place $n$ disks in an image, each of which can have a different position and radial size. In cases of $n>1$ disks, the disks are allowed to overlap with one another, but not exceed the boundaries of the image. For $n>1$, the dataset also introduces the additional complexity that the ideal embedding is non-unique, i.e. the interchange of position and size parameters between two disks will result in the same image. Examples of samples from this generator are shown in Figures \ref{fig:ex_disk} and \ref{fig:compress_hetero}. An explicit algorithm for this generator is described in Algorithm \ref{alg:ndisk} in Appendix \ref{apx:implementation}.

We generate four separate $N=10,000$ datasets for each case of 1-, 2-, 3-, and 4-Disk images. We also construct a heterogenous dataset of $N=10,000$ generated images in which the number of disks is uniformly sampled from $n\in\{1,2,3,4\}$. This latter set introduces the interesting case in which two disks may entirely overlap, and thus reduce the effective $n$ by one. This behavior is explored in Section \ref{sec:exploration}. When encoding this dataset, we use a feed-forward autoencoding convolutional neural network architecture with three convolutional layers and three dense layers in each of the encoder $e_\phi$ and decoder $d_\eta$. The maximum width of the bottleneck is $\kmax=16$.

\subsection{CLIP embeddings of MS-COCO}
Lastly, we demonstrate the ability of IOBs to compress latent representations of SOTA zero-shot image-to-image and text-to-image models. Specifically, we use IOBs to compress the distribution of the MS-COCO 2017 Captioning dataset \cite{lin2014microsoft} in the Contrastive Language-Image Pre-Training (CLIP) embedding space \cite{radford2021learning}. In this fashion, we seek to study how the IOBs handle, compress, and order information in semantic embeddings from text and image data. Transforming the full suite of MS-COCO images into CLIP embeddings produces $\sim100,000$ datapoints in $\bfx\in\bbR^{768}$. We then use a deep feed-forward autoencoder with three-layer encoders and decoders with a maximum bottleneck width of $\kmax=384$. Several examples of MS-COCO prompts and images are shown in Figure \ref{fig:examples}, and more are given in Figures \ref{fig:ex_coco_img_apx} and \ref{fig:ex_coco_txt_apx} in Appendix \ref{apx:supplement}.

We evaluate the compression quality of our IOB autoencoders using the image-to-image and text-to-image pipelines from unCLIP \cite{ramesh2022hierarchical}. unCLIP uses a conditional diffusion model to generate sample images from CLIP image embeddings. It also includes a diffusion prior which translates text prompts into CLIP image embeddings, which are then used to condition the aforementioned image generators. In our tests, we transform MS-COCO images and captions into the CLIP semantic space using the pre-trained CLIP encoder and unCLIP diffusion prior, compress and reconstruct them with our latent autoencoders, and use the outputs to condition the unCLIP generative image model to create samples which look semantically similar to the MS-COCO prompts and images. We use the standard unCLIP diffusion prior to map prompts to CLIP embeddings and a finetuned version of Stable Diffusion 2.1 \cite{Rombach_2022_CVPR} for generating images.



\begin{figure}[!t]
    \centering
    \begin{subfigure}[t]{0.4\textwidth}
        \centering
        \includegraphics[width=\textwidth]{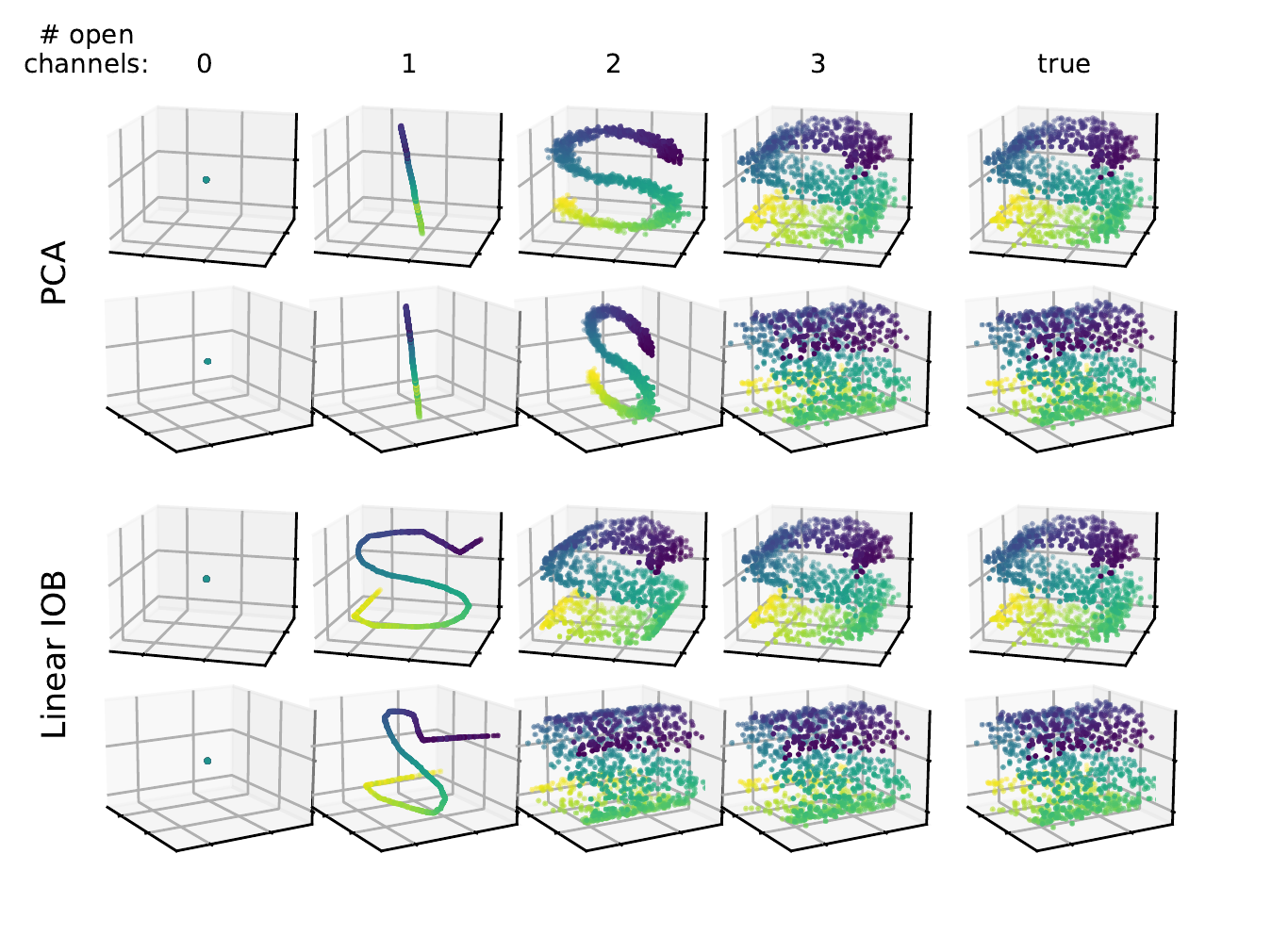}
        \caption{S-curve}
        \label{fig:ex_s}
    \end{subfigure}\hfill
    \begin{subfigure}[t]{0.57\textwidth}
        \centering
        \includegraphics[width=\textwidth]{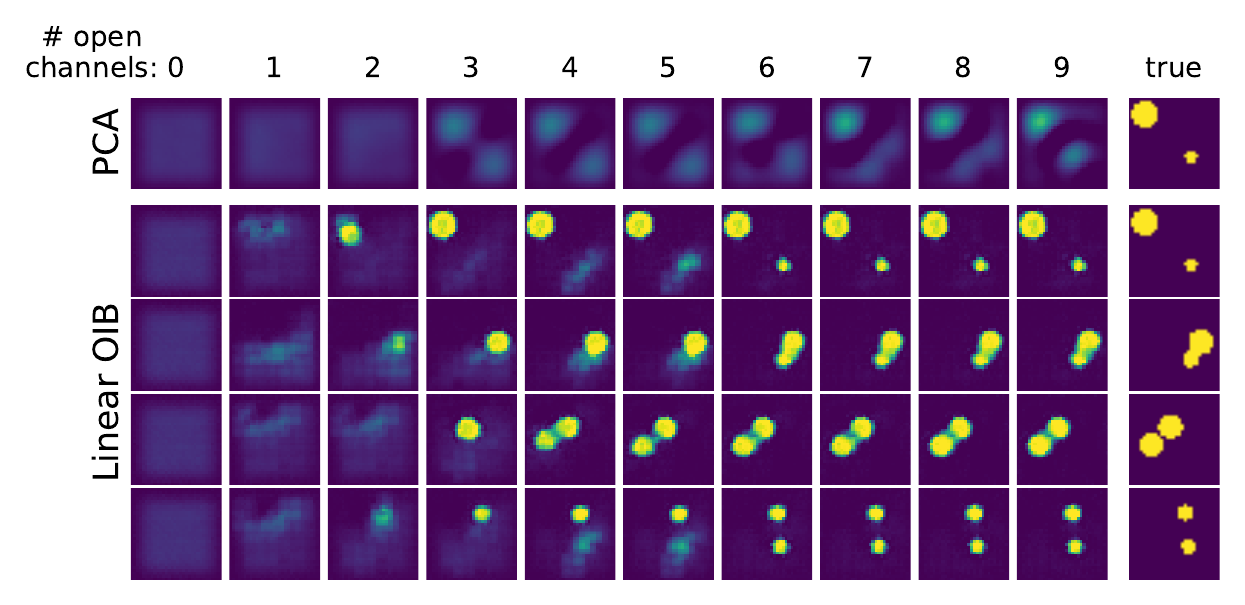}
        \caption{$2$-Disk}
        \label{fig:ex_disk}
    \end{subfigure}\\
    \begin{subfigure}[b]{\textwidth}
        \centering
        \includegraphics[width=0.8\textwidth]{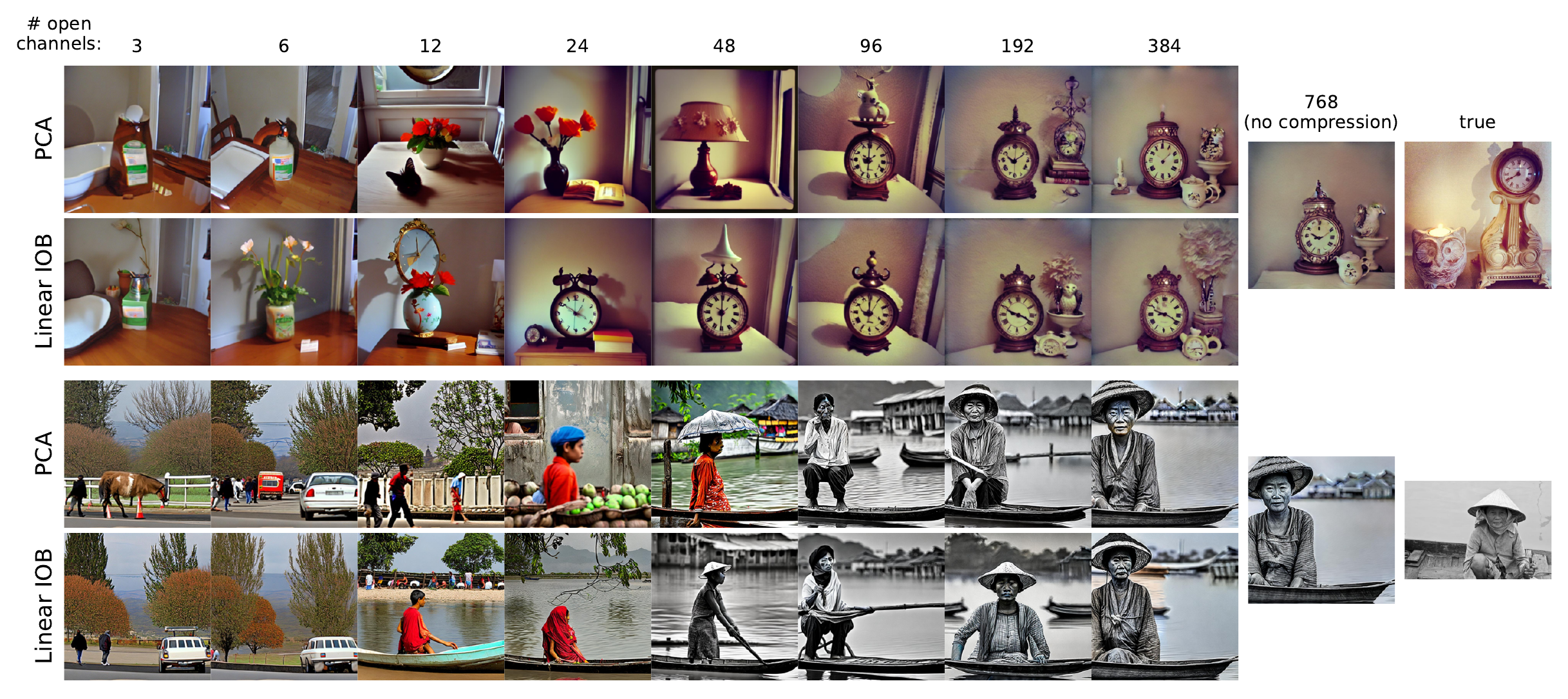}
        \caption{MS-COCO image-to-image}
        \label{fig:ex_unclip_img}
    \end{subfigure}\\
    \begin{subfigure}[b]{\textwidth}
        \centering
        \includegraphics[width=0.8\textwidth]{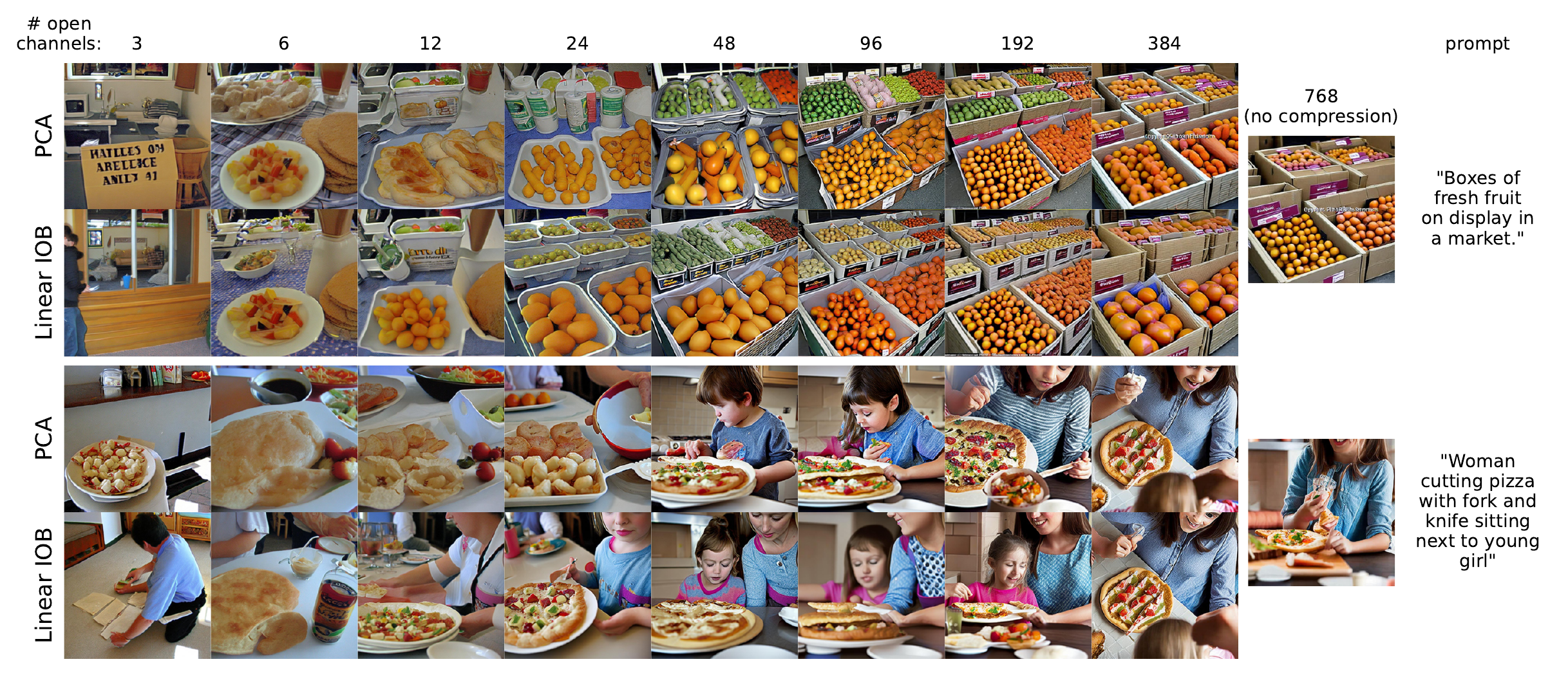}
        \caption{MS-COCO text-to-image}
        \label{fig:ex_unclip_txt}
    \end{subfigure}
    \caption{Reconstruction examples of various experiments as a function of bottleneck width after autoencoding compression with either PCA or Linear IOBs. For the MS-COCO experiments, images are generated using the same diffusion noise for all bottleneck widths, both for PCA and Linear IOB examples.}\label{fig:examples}
\end{figure}

\section{Compression}\label{sec:compression}
In this section, we empirically test the ability of IOBs to adaptively compress the datasets described in Section \ref{sec:experiments}. We quantitatively and qualitatively compare these compressions to those of several traditional and machine learning baselines for ordered encoding. We observe that our IOB procedure naturally captures and orders semantic information and compresses latent spaces more efficiently than all of our tested baselines.

\subsection{Baselines}
Our primary baseline is Principle Component Analysis (PCA) \cite{jolliffe2016principal}, a general linear embedding strategy for high-dimensional compression. PCA works by finding an orthogonal linear projection which transforms the data vector to a new coordinate system that maximizes the data variance at each cardinal axis. PCA fails in cases where features are non-linearly related, resulting in extremely inefficient compression.


We also train a suite of normal autoencoders that account for every possible bottleneck width. These serve as a benchmark for the compression ability of our autoencoder architectures when we relax the disentanglement constraint and do not enforce an adaptive ordered bottleneck. Without these constraints, the separate normal autoencoders should be able to the explain more variability in our dataset than those with IOBs, for a given bottleneck size. However, we note that it is still possible for the reverse to be true in some cases, and we observe and explain such behavior in the subsequent section.

\subsection{Results}

\begin{figure}[!t]
  \centering
  \begin{subfigure}[t]{0.329\textwidth}
      \centering
      \includegraphics[width=\textwidth]{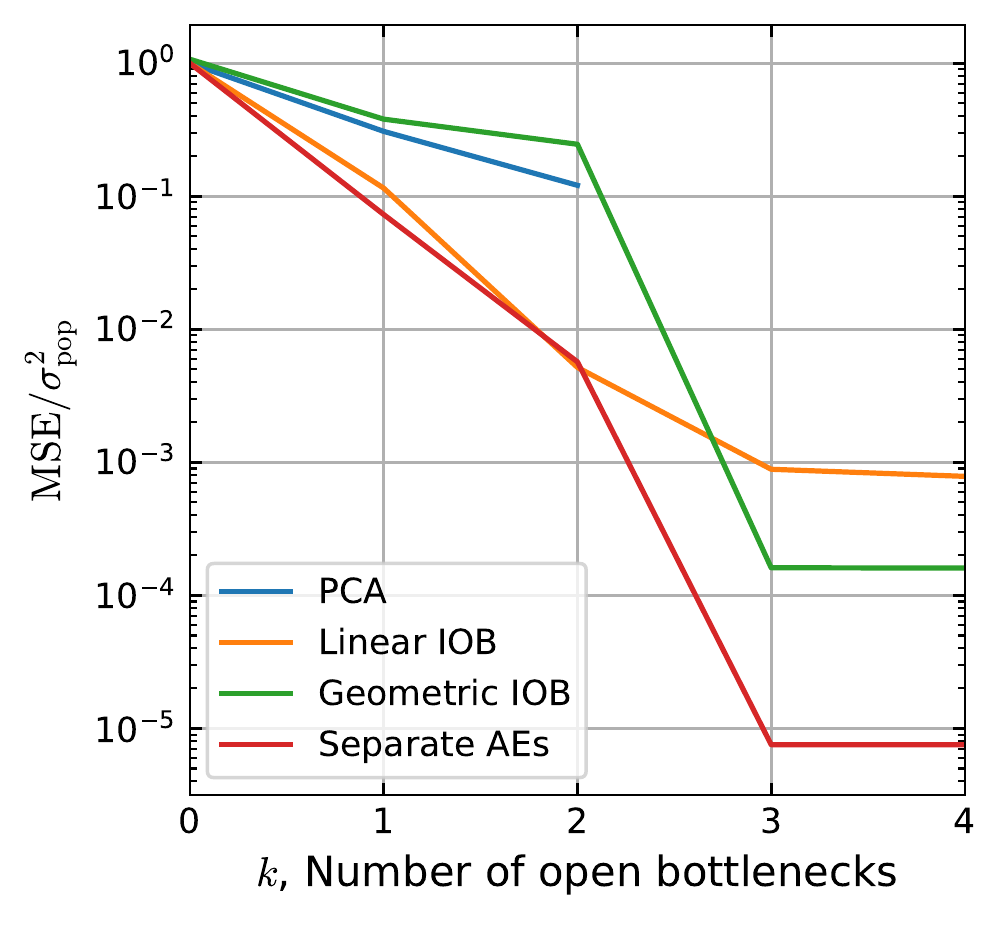}
      \caption{S-curve}
      \label{fig:compress_s}
  \end{subfigure}
  \begin{subfigure}[t]{0.329\textwidth}
      \centering
      \includegraphics[width=\textwidth]{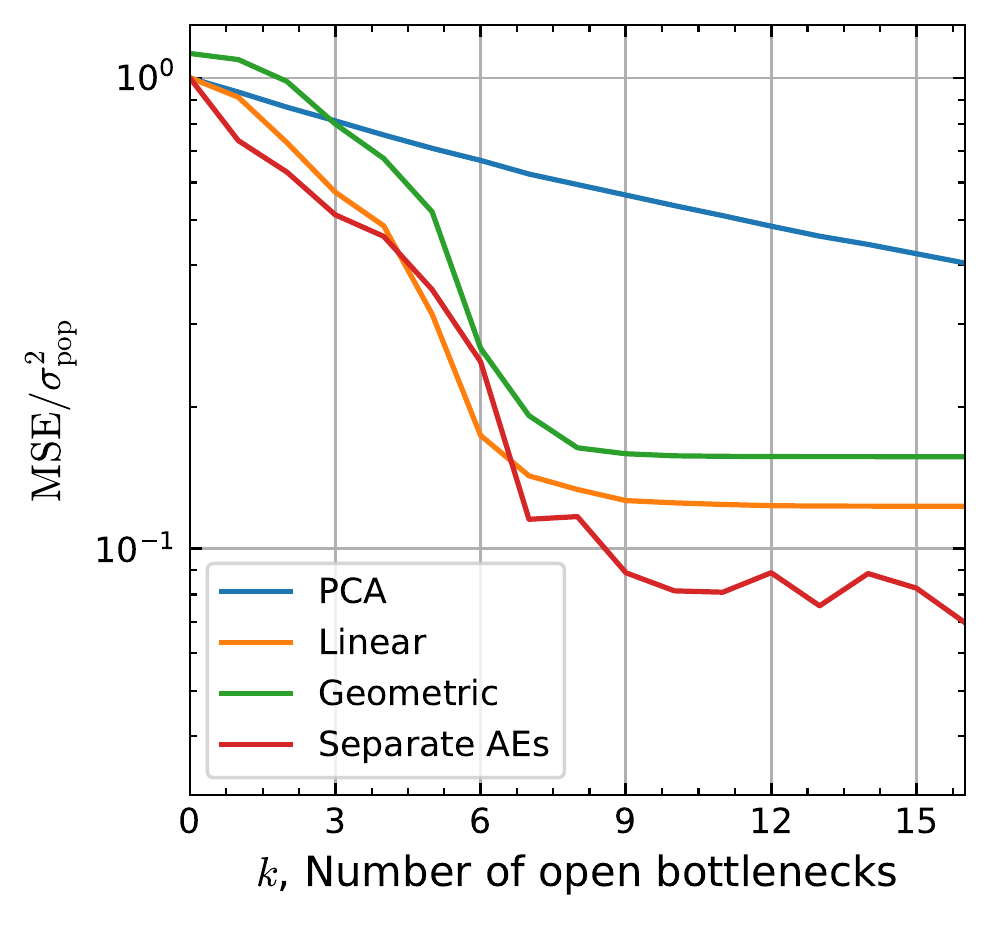}
      \caption{$2$-Disk}
      \label{fig:compress_ball}
  \end{subfigure}
  \begin{subfigure}[t]{0.329\textwidth}
      \centering
      \includegraphics[width=\textwidth]{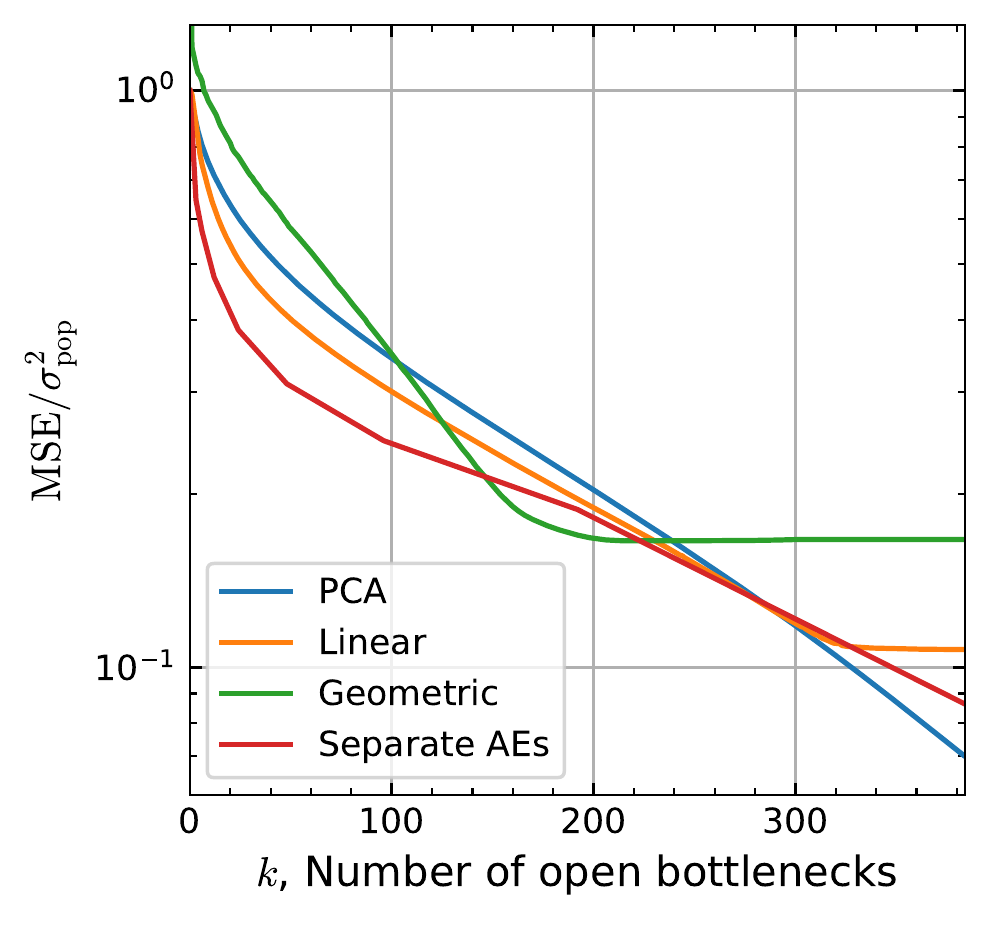}
      \caption{MS-COCO}
      \label{fig:compress_coco}
  \end{subfigure}
  \caption{Average mean-squared-error (MSE) reconstruction loss for different autoencoding compression schemes. All MSE's are normalized by the fixed population variance. We note that for the S-curve dataset, the PCA loss goes to 0 at $k=3$, and so is not shown in its plot. }\label{fig:compression}
\end{figure}

Figure \ref{fig:compression} demonstrates the reconstruction performance of our Linear IOB model on the S-curve, 2-Disk, and MS-COCO experiments. The reconstructions are plotted relative to an equivalent reconstruction with PCA, as the number of open bottlenecks at inference time increases. We observe a clear growth in information content as the bottleneck widens. Qualitatively, the reconstructions exhibit semantic improvements, with underlying concepts gradually appearing as sufficient information is available. In the 2-Disk example, the reconstructions reveal one disk at a time, completing at around $k=3$ and $k=6$. This contrasts with PCA, which approaches the truth through the mean field reconstruction. Similarly, in the MS-COCO experiment, large-scale features are reconstructed first (e.g., object types, camera conditions) followed by smaller, more specific features (e.g., backgrounds, clothing, secondary objects) at higher bottleneck widths. It's important to note that the embedding IOB models used in the MS-COCO example were trained with MSE minimization of CLIP latents and evaluated with a fixed diffusion noise schedule. Therefore, they can recover semantic features of the fully-open unCLIP model (768 channels in Figure \ref{fig:compress_coco}) but not exactly those of the true image.

In autoencoder experiments, the learned representations are akin to $k$-dimensional mainfold-fitting. In the S-curve reconstruction, we first see the reconstructions falling on the mean point at $k=0$, then tracing a the S-line at $k=1$, and finally matching the unbiased S-curve manifold (without noise) at $k=2$. It is only when the problem is full rank (at $k=3$) that the model attempts to recover noise properties in the full $\bbR^3$ space. 
Here, we note that although a sufficiently flexible neural network could represent the entire S-curve in a single latent (using an infinitely long line tracing the manifold shape), the IOBs here converge to the simpler solution of fitting a $k$-dimensional manifold as an approximation to the dataset.

Figure \ref{fig:compression} shows the reconstruction performance on the full test set, plotting the mean reconstruction error as a function of bottleneck width for each of our IOB autoencoders and benchmark compressors. As expected, PCA performs extremely poorly on most non-linear datasets, except in the case of high-$k$ in the MS-COCO, wherein small variations in CLIP embeddings are poorly fit by the autoencoders. Secondly, we observe that Linear IOBs provide better compression at almost all bottleneck widths than Geometric IOBs. We provide two compounding reasons for this behavior. First is that, when implementing Equation \ref{eqn:objective}, the stochastic summation is considerably less constraining than when it can be evaluated at full width. Second, the `unit sweeping' method of training the Geometric IOB solves its tasked problem, i.e. preventing vanishing gradients, but also causes the latent representations learned at the beginning of training to drift by the end of training. This gives rise to the interesting behavior in the MS-COCO compression, wherein we have optimized for compression at bottleneck widths $k\geq 200$, while sacrificing compression performance at widths $k<10$. 

The separate autoencoders serve as our reference for `optimal training' of each assumed architecture, while relaxing any ordering or disentanglement constraints. However, we find that there are several cases in each experiment in which the Normal IOB or the Geometric IOB outperforms the separate autoencoders for some $k$. This is likely the result of stochasticity in the initialization and training procedure or inductive bias learned through the IOB loss. In any case, in this apples-to-apples comparison, we find that the best IOB models approach or exceed the compression performance of these optimal autoencoders.


Lastly, we observe that reconstructions of the IOB models and separate autoencoders saturate at or around the intrinsic dimensionality of each dataset. For example, the S-curve reconstructions very quickly approach $1\%$ error at $k=2$, the dimensionality of the S-manifold, and those of the 2-Disk experiment asymptote with $10\%$ reconstruction error at $k=6 = 3n$ where $n=2$ disks. The fact that these models do not reach error $0$ is a natural limitation of our finite network architectures. This saturation is used to demonstrate Intrinsic Dimensionality estimates in Section \ref{sec:dimensionality}.

\section{Global Intrinsic Dimensionality Estimation} \label{sec:dimensionality}
Given the unique structure of the IOB, we can perform nested model comparison at inference time to compare different bottleneck widths and estimate intrinsic dimensionality. First, let us briefly expand on the notation from Section \ref{sec:theory}. The IOB layer $b_k:\bbZ\rightarrow\bbZ$ creates a bottleneck by masking the outputs of the $i$-th latent embeddings above $i>k$. We can consider this an element-wise multiplication, i.e. $g_k(\bfz) := \bfz \circ \bfe_k$, between the input latents $\bfz$ and a masking vector $\bfe_k := \left(\mathbbm{1}[k\geq1], \mathbbm{1}[k\geq2],\dots, \mathbbm{1}[k\geq\kmax]\right)\in [0,1]^{\kmax}$, where $\mathbbm{1}$ is the indicator function. Next, we assume we have found optimal parameters $\theta^* = (\phi^*, \eta^*)$ following from Equation \ref{eqn:objective}, allowing us to use our trained encoder $e_{\phi^*}$ to compress our test set inputs from $\mathcal{D}_\mathrm{test} = \{\bfx_i, \bfy_i\}_{i=1}^{N_\mathrm{test}}$  into a set of latents $\{\bfz_i\}_{i=1}^{N_\mathrm{test}}$. Then, we define a composite log-likelihood, $\ell_{\bbZ} : \bbZ \rightarrow \bbR$, which takes as input the post-bottleneck latents and produces the log-likelihood of the test set in reconstruction space.
\begin{equation}
    \ell_{\bbZ} (\bfz, \bfy) = \ell \left[d_{\eta^*}(\bfz), \bfy\right].
\end{equation}
In essence, what we have done here is reframe our trained encoder-decoder network to evaluate the log-likelihood of test data \textit{in latent space}. Our goal now is to derive an estimator which maximizes the composite log-likelihood $\ell_{\bbZ}$ when given the pre-trained encoder's embeddings. Our estimator of choice is $g(\bfz;\bfe) =\bfz\circ\bfe$, a generalization of the linear $g_k$ bottleneck operation, but now with masking parameters $\bfe$ which can vary anywhere in real space $\bbR^{\kmax}$. Due to the training procedure of $d_{\eta^*}$,  we would expect that this $\bfe^* = \mathbf{1}$, where $\mathbf{1}$ is a vector of all ones. However, we can use the flexibility of these new variable parameters to quantify the improvement in $\ell_{\bbZ}$ as we allow more $\bfe$ components to vary.

We perform a likelihood ratio test to quantify the statistical significance of increasing the bottleneck width on the test data likelihood. Let us consider a null hypothesis $H_0$ in which $k$ bottleneck connections are open and compare it to an alternative hypothesis $H_1$ with $k+1$ connections open. We can frame this explicitly in terms of the mask parameters $\bfe$, such that $H_0: e_i = 0\ \forall\ i>k$ and $H_1: e_i = 0\ \forall\ i>k+1$. Following from Wilk's theorem \cite{wilks1938large}, we expect that the distribution of twice the difference in the log-likelihood,
\begin{equation}
    D = 2\left(\ell_{\bbZ}(H_1) - \ell_{\bbZ}(H_0)\right),
\end{equation}
will asymptote to a $\chi^2$ distribution for sufficient test data. Here, we use the notation $\ell_{\bbZ}(H_i)$ to represent the maximum log-likelihood $\ell_{\bbZ}$ averaged over the whole test set under hypothesis $H_i$. Using this test, we can examine these hypotheses down to a specified tolerance level (in our case $\alpha=0.05$) to determine whether we need $k+1$ parameters to describe the data. We then assign an intrinsic dimensionality to our dataset equal to the bottleneck width when we cannot reject the null hypothesis, i.e when we do not see sufficient gain in the log-likelihood for increasing $k$. We note that this method gives us an estimate of the global intrinsic dimensionality, integrated over all points in our test set, but could be extended to analysis of local intrinsic dimensionality in future work.


Using the implementations in \cite{bac2021scikit}, we compare our recovered intrinsic dimensionality estimates against a series of baseline models from the literature. A full list of tested baselines is shown in Table \ref{tab:dimension}. These include several algorithms based in linear compression (such as PCA) \cite{cangelosi2007component}, nearest-neighbor finding \cite{facco2017estimating}, and manifold-fitting \cite{farahmand2007manifold}. 

Table \ref{tab:dimension} compares the dimensionality estimates of our methods with the baselines, true dimensionality, and data dimensionality. The performance of the estimators varies across the S-curve, $n$-Disk, and MS-COCO CLIP embedding datasets due to their different characteristics. While all estimators perform well on the S-curve dataset, differences arise in the $n$-Disk dataset, where PCA, MADA, and TwoNN estimate higher dimensionalities compared to the true dimensionality. However, IOB models provide closer estimates, often predicting to within one dimension of the truth. The IOB models generally overestimate the intrinsic dimensionality, possibly due to insufficient fitting or non-unique embeddings in the $n$-Disk datasets (i.e. when multiple balls either overlap or swap positions).
Lastly, whereas the true dimensionality of MS-COCO embeddings in the CLIP semantic embedding space is unknown, we include intrinsic dimensionality estimates for completeness. Interestingly, the Linear IOB estimates a intrinsic dimensionality of $322$, which is very close to the $319$-dimensional PCA compression that the authors of \cite{ramesh2022hierarchical} used to construct the unCLIP prior. We suggest an exhaustive comparison of the CLIP embedding dimensionalities for future work.

\begin{table}[!t]
    \centering
    \begin{tabular}{lllllll}
    \toprule
     ID Estimator & S-curve & $1$-Ball & $2$-Ball & $3$-Ball & $4$-Ball & MS-COCO CLIP\\
     \midrule
     PCA \cite{cangelosi2007component} & 3 & 33 & 37 & 39 & 38 & 106 \\
     MADA \cite{farahmand2007manifold} & 2.5 & inf & 13.2 & 16.9 & 19.5 & 22.7 \\
     TwoNN \cite{facco2017estimating} & 2.9 & 5.3 & 13.6 & 16.3 & 21.4 & 21.4 \\
     Linear IOB* & \textbf{2} & \textbf{3} & \textbf{7} & \textbf{10} & 14 & 322 \\
     Geometric IOB* & 3 & \textbf{3} & \textbf{7} & \textbf{10} & \textbf{12} & 196\\
     \midrule
     Data Dimensionality & 3 & 1024 & 1024 & 1024 & 1024 & 768 \\
     True Dimensionality & 2 & 3 & 6 & 9 & 12 & $\leq768$ \\
     \bottomrule
    \end{tabular}
    \caption{Global intrinsic dimensionality estimates for various estimators on synthetic and real datasets. Data dimensionality refers to the the dimensionality $\dim(\bbX)$ of each sample in the training dataset, wheras true dimensionality refers to the number of tunable parameters used in the forward model of the dataset generation. Models introduced in this work are marked with an asterisk.}
    \label{tab:dimension}
\end{table}

\section{Data Exploration} \label{sec:exploration}

\begin{figure}
  \centering
  \begin{subfigure}[t]{0.12\textwidth}
      \centering
      \includegraphics[width=0.8\textwidth]{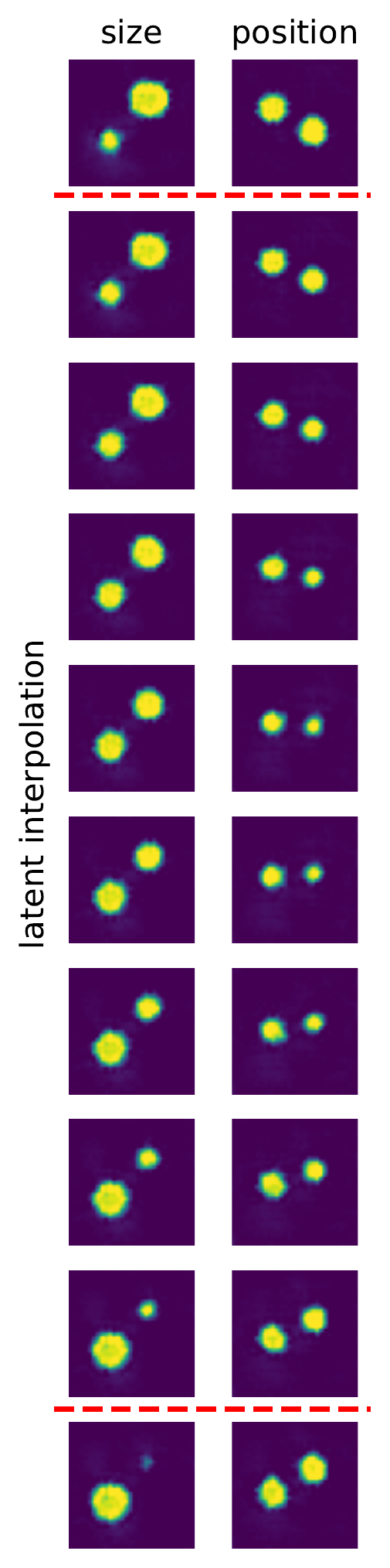}
      \caption{Interpolation}
      \label{fig:interp}
  \end{subfigure}
  \hfill
  \begin{subfigure}[t]{0.87\textwidth}
    \centering
    \includegraphics[width=0.9\textwidth]{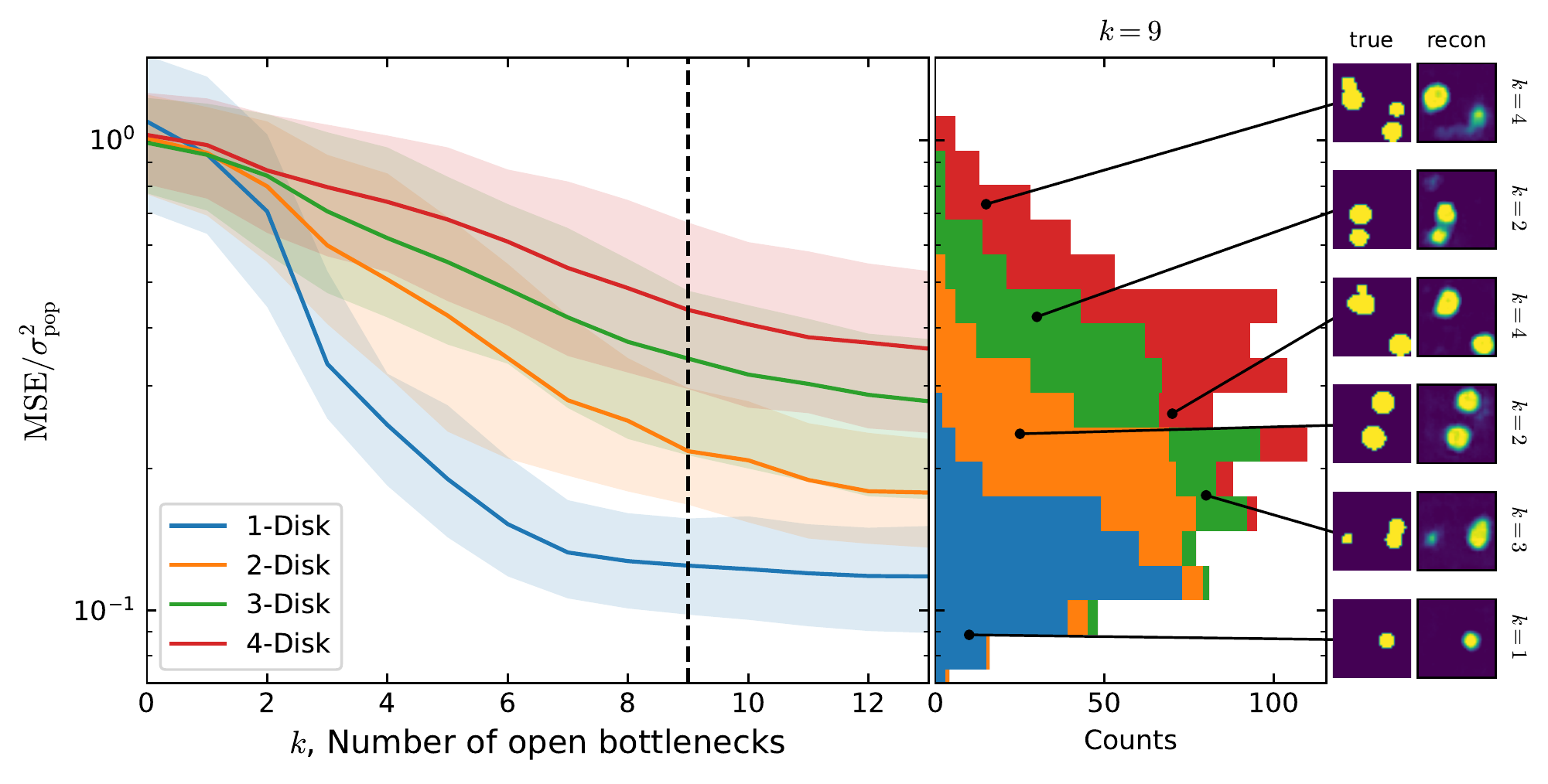}
    \caption{Heterogenous compression}
    \label{fig:compress_hetero}
\end{subfigure}
  \caption{Data exploration techniques with IOBs. In \textbf{(a)}, reconstructions derived from encoding and decoding true samples are shown above the upper and below the lower red lines. All images between the red lines are derived from interpolating between these true embeddings in the IOB latent space and decoding them into image space. In \textbf{(b)}, we show the median and 16-84th percentile intervals of each $n$-Disk population in a heterogenous compression. Example reconstructions on the right are drawn from the reconstructions at a bottleneck width of $k=9$, and are shown with with a black line corresponding to their position in the distribution and their true number of disks $n$ at the right.}
\end{figure}
In Figure \ref{fig:interp}, we demonstrate an example of interpolation within the latent space of a Linear IOB autoencoder trained on the $2$-Disk dataset. By performing simple linear interpolation of the IOB latent variables, we can observe variations in the size or location of disks, resulting in the generation of novel images not present in the original training set. This suggests that the latent variables potentially encode semantic information related to the parameters used for generating the $2$-Disk samples. Such interpolation provides insights into the capabilities of the IOB model in capturing and manipulating underlying features within the data.

Figure \ref{fig:compress_hetero} shows how training on a heterogenous dataset of mixed $n$-Disk examples from $n\in\{1,2,3,4\}$ can lead to complexity rank-ordering and pattern discovery. At inference time, we clearly see a structured ordering of different image complexities with simpler images receiving lower MSE at inference time. Clearly, the least complex $1$-Disk samples achieve optimal compression at earlier bottleneck widths, whereas more complex samples such as $3$- or $4$-Disks show much poorer compression under smaller bottlenecks. However, this scaling is not monotonic and we see considerable mixing among the $n$-Disk populations. This is the result of the behavior of disks in our images to partially or entirely overlap, reducing the effective number of disks $n$ by one. We demonstrate some of such examples at the right of Figure \ref{fig:compress_hetero}.



\section{Conclusions} \label{sec:conclusions}

In conclusion, this paper introduced a method for ordering latent variables based on their impact in minimizing the likelihood, offering a unified framework that incorporates multiple previous approaches. The results demonstrate the adaptive compression capabilities of the proposed method, achieving near-optimal data reconstruction within a given neural architecture while capturing semantic features in the inferred latent ordering. Additionally, the model proves effective in compressing high-dimensional data when used in conjunction with SOTA multimodal image-to-image and text-to-image models like unCLIP\cite{ramesh2022hierarchical}. Furthermore, a novel methodology for intrinsic dimensionality estimation is introduced, surpassing previous benchmarks through various synthetic experiments. For further details on limitations, please refer to Appendix \ref{apx:limit}. 


\section{Acknowledgements}
MH is the corresponding author. MH is supported by the Simons Collaboration on Learning the Universe. The Flatiron Institute is supported by the Simons Foundation.


\bibliographystyle{unsrt}
\bibliography{bibliography}

\newpage

\appendix

\section{Limitations} \label{apx:limit}

In the proposed methods for intrinsic dimensionality estimation, we describe statistical tests of likelihood gain within the latent space of a pre-trained encoder-decoder. In reality, the intrinsic dimensionality we recover is highlydependent on the architecture and training limitations of the implemented neural networks. In this way, our chosen architectures act as hyperparameters to our dimensionality estimates. Although we expect that increasing the depth of the network would allow our dimensionality estimates to asymptote to some value, we have not explicitly tested this.

In addition, the unit sweeping procedure used to implement the geometric bottleneck does not produce an optimal fitting across all possible bottleneck widths and is dependent on the convergence criterium and the choice of geometric rate $r$. In our applications, we tried a several of rates $r$ which produced stable training, but did not fully explore the space of possible hyperparameters. We expect this to have an impact on the learned compressions as well as the recovered estimates of intrinsic dimensionality.


\section{Supplementary Examples} \label{apx:supplement}

\begin{figure}[!htb]
    \centering
    \includegraphics[width=\textwidth]{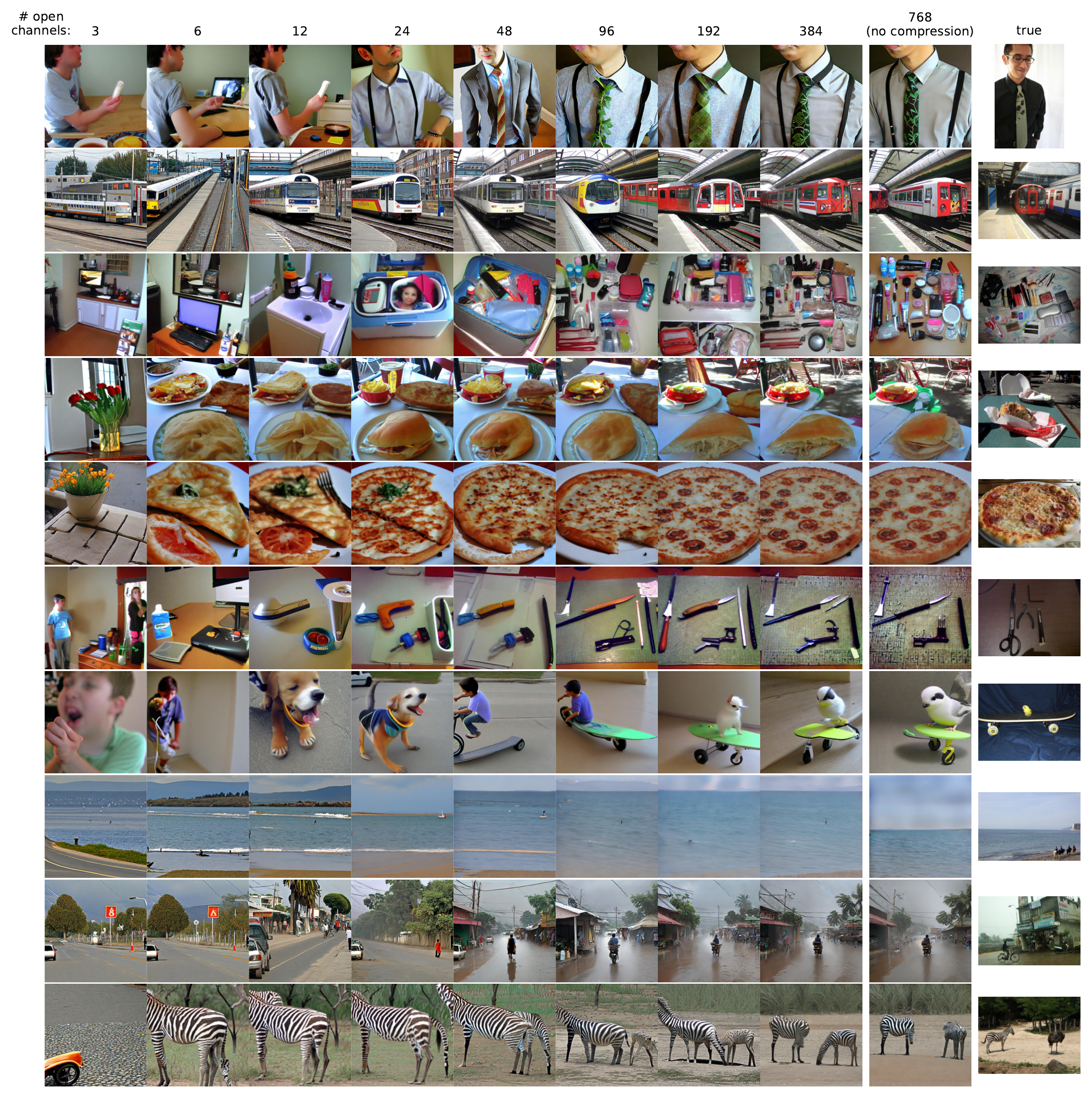}
    \caption{Supplementary examples of the MS-COCO image-to-image generation using latent compression with the Linear IOBs. Reconstructions are shown as a function of bottleneck width. Images are generated using the same diffusion noise for all bottleneck widths.}
    \label{fig:ex_coco_img_apx}
\end{figure}
\begin{figure}[!htb]
    \centering
    \includegraphics[width=\textwidth]{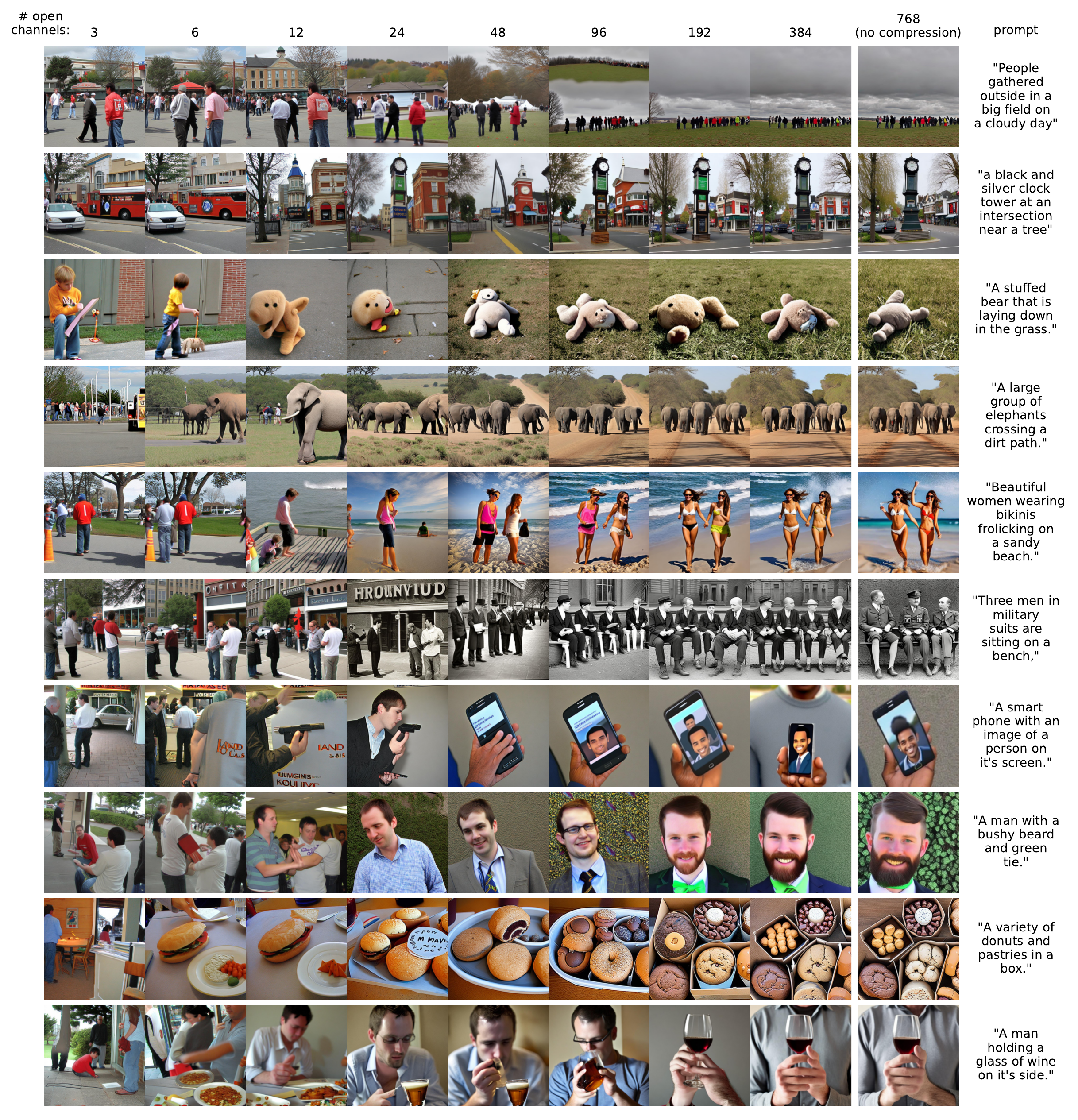}
    \caption{Supplementary examples of the MS-COCO text-to-image generation using latent compression with the Linear IOBs. Reconstructions are shown as a function of bottleneck width. Images are generated using the same diffusion noise for all bottleneck widths.}
    \label{fig:ex_coco_txt_apx}
\end{figure}

\section{Implementation details} \label{apx:implementation}

\subsection{Training procedure}
All datasets undergo a $90\%-10\%$ training-validation split which are fixed for all experiments. All figures and results shown in the manuscript are evaluated solely on the validation split. In all experiments, we use an Adam optimizer \cite{kingma2014adam} with a learning rate of $5\times 10^{-5}$, no learning rate decay, and a batch size of 64. For all models except the Geometric IOB, we implement an early stopping criterion of a minimum of $0.01\%$ validation improvement over $20$ epochs. For the Geometric IOB, we perform unit sweeping over every latent node, wherein our criterion for `convergence' of each node is a minimum of $1\%$ validation improvement over $10$ epochs. All neural network models are implemented in \texttt{Pytorch} \cite{paszke2019pytorch}. 

All experiments use a Gaussian log-likelihood with fixed variance for construction of the loss function in Equation \ref{eqn:objective}. The varaince is set to be the population variance, estimated empirically from the training set. In the case of the heterogenous compression for Figure \ref{fig:compress_hetero}, $n$-Disk samples are 

\subsection{Experiments}
The maximum width of bottleneck for each experiment were chosen to avoid computational intractability while maximizing the expected information gain in the bottleneck range. The architectures were chosen to mimic examples of previous deep fully-connected and convolutional autoencoders. However, we did not perform an exhaustive architecture search and leave this to future work.

For the S-curve dataset, we use fully-connected dense encoders and decoders. Including the input layer and the bottleneck, the sequence of nodes in each layer of the encoder is $3-64-64-4$. The sequence of decoder layers is simply the reverse of that of the encoder. All layers use a ReLU activation function. For the Geometric IOB of S-curve dataset, we use a rate of $r=0.95$.

The generator for the $n$-Disk dataset is described explicitly in Algorithm \ref{alg:ndisk}. The autoencoder for this experiment is a convolutional neural network. The encoder $e_\phi$ is constructed from three convolution layers with respectively $4$, $12$, and $24$ filters each of size $4\times4$, a stride of two, and edge padding of one. The convolutional layers are followed by three dense layers of width $256-128-16$, including the bottleneck of size $\kmax=16$. The decoder $d_\eta$ has the reversed architecture of the encoder, except it uses convolutional upsampling layers instead of normal convolutional layers. All layers in both the encoder and decoder use an ReLU activation function. For the Geometric IOB of all $n$-Disk datasets, we use a rate of $r=1/3$.

Lastly, we use fully-connected dense encoders and decoders for the compression of CLIP embeddings of MS-COCO images. Including the input layer and the bottleneck, the sequence of nodes in each layer of the encoder is $768-384-384-384$. As in the S-curve experriment, the sequence of decoder layers is simply the reverse of that of the encoder and we use ReLU activation for each layer. For the Geometric IOB of the MS-COCO dataset, we use a rate of $r=0.05$.

\begin{algorithm}
  \SetKwInOut{Input}{Input}
  \SetKwInOut{Output}{Output}
  \Input{Number of disks $n$}
  \Output{A single-channel image $\bfx$ of size $32\times32$ wherein $x_{ij}\in[0,1]\ \forall\ i,j$}
  \For{$i \gets 1$ \textbf{to} $n$} {
    $r_i \sim \mathcal{U}(2,5)$ \\
    $a_i \sim \mathcal{U}(r_i, 32-r_i)$\\
    $b_i \sim \mathcal{U}(r_i, 32-r_i)$\\
    \For{$j \gets 1$ \textbf{to} $32$} {
      \For{$k \gets 1$ \textbf{to} $32$} {
        \uIf{$r_i^2 \geq (a_i - j+0.5)^2 + (b_i-k+0.5)^2$} {
          $x_{jk}\gets 1$\;
        }
        \Else{
          $x_{jk} \gets 0$\;
        }
        }
    }
  }
  \Return{$\bfx$}
  \caption{Synthetic generation of an image in the $n$-Disk datasets.}\label{alg:ndisk}
\end{algorithm}

\end{document}